\documentclass{bmvc2k}

\title{Axial-Centric Cross-Plane Attention for\\ 3D Medical Image Classification}

\addauthor{Doyoung Park}{park.doyoung@nhcs.com.sg}{1,2}
\addauthor{Jinsoo Kim}{tjdujd1212@gmail.com}{3}
\addauthor{Lohendran Baskaran}{lohendran_baskaran@duke-nus.edu.sg}{1,2,$^{\star}$}

\addinstitution{
 National Heart Centre Singapore,\\
 Singapore
}
\addinstitution{
 CVS.AI,\\
 National Heart Research Institute of Singapore,\\
 Singapore
}
\addinstitution{
 Independent Researcher,\\
 Republic of Korea
}

\runninghead{Park, Kim, Baskaran}{Axial-Centric Cross-Plane Attention}

\def\etal{\emph{et al}\bmvaOneDot}

\begin{document}

\maketitle

\begingroup
\renewcommand\thefootnote{}
\footnotetext{$^{\star}$ Corresponding author}
\endgroup

\begin{abstract}
Clinicians commonly interpret three-dimensional (3D) medical images, such as computed tomography (CT) scans, by examining multiple anatomical planes rather than relying solely on volumetric representations. In this multi-planar approach, the axial plane often serves as the primary acquisition and diagnostic reference in many CT interpretation workflows, while the coronal and sagittal planes provide complementary spatial information to increase diagnostic confidence. However, many existing 3D deep learning methods either process volumetric data holistically or assign equal importance to all planes, failing to reflect this asymmetric and axial-centric clinical interpretation workflow. To address this limitation, we propose an axial-centric cross-plane attention architecture for 3D medical image classification that captures the inherent asymmetric dependencies between different anatomical planes. Our architecture incorporates MedDINOv3, a medical vision foundation model pretrained via self-supervised learning on large-scale axial CT images, as a frozen feature extractor for the axial, coronal, and sagittal planes. RICA blocks and intra-plane transformer encoders capture plane-specific positional and contextual information within each anatomical plane, while axial-centric cross-plane transformer encoders selectively condition axial features on complementary information from auxiliary planes. Experimental results on six datasets from the MedMNIST3D benchmark demonstrate that the proposed architecture consistently outperforms existing 3D and multi-plane models in terms of accuracy and AUC. Notably, a reduced-capacity variant (AC-Tiny) achieves highly competitive performance with significantly fewer trainable parameters, indicating that architectural design plays a more crucial role in performance improvements than increasing model scale. Extensive ablation studies further confirm the importance of axial-centric querying and query-key-value allocation, directional cross-plane fusion, residual-free cross-attention, and classification head design. Additionally, slice-level Grad-CAM-based feature attribution visualizations demonstrate that the proposed architecture identifies diagnostically relevant anatomical regions across all anatomical planes. These findings highlight the importance of aligning architectural design with clinical interpretation workflows for robust 3D medical image analysis under limited-data settings.
\end{abstract}

\section{Introduction}
\label{sec:intro}
In clinical practice, three-dimensional (3D) medical images, such as computed tomography (CT) scans, are commonly interpreted using multiple anatomical planes rather than as a single volumetric representation. In the multi-planar approach, the axial plane serves as the primary acquisition and diagnostic plane in many CT interpretation workflows, while the coronal and sagittal planes are reconstructed using multiplanar reformation (MPR) to enhance visualization of spatially complex anatomical structures, improving diagnostic accuracy and clinical confidence~\cite{num1,num2}. Clinicians primarily detect lesions on axial planes and subsequently examine auxiliary planes in a structured manner, reflecting inherently asymmetric dependencies among anatomical planes during clinical decision-making.

\begin{figure}
\begin{center}
\end{center}
\includegraphics[width=\textwidth]{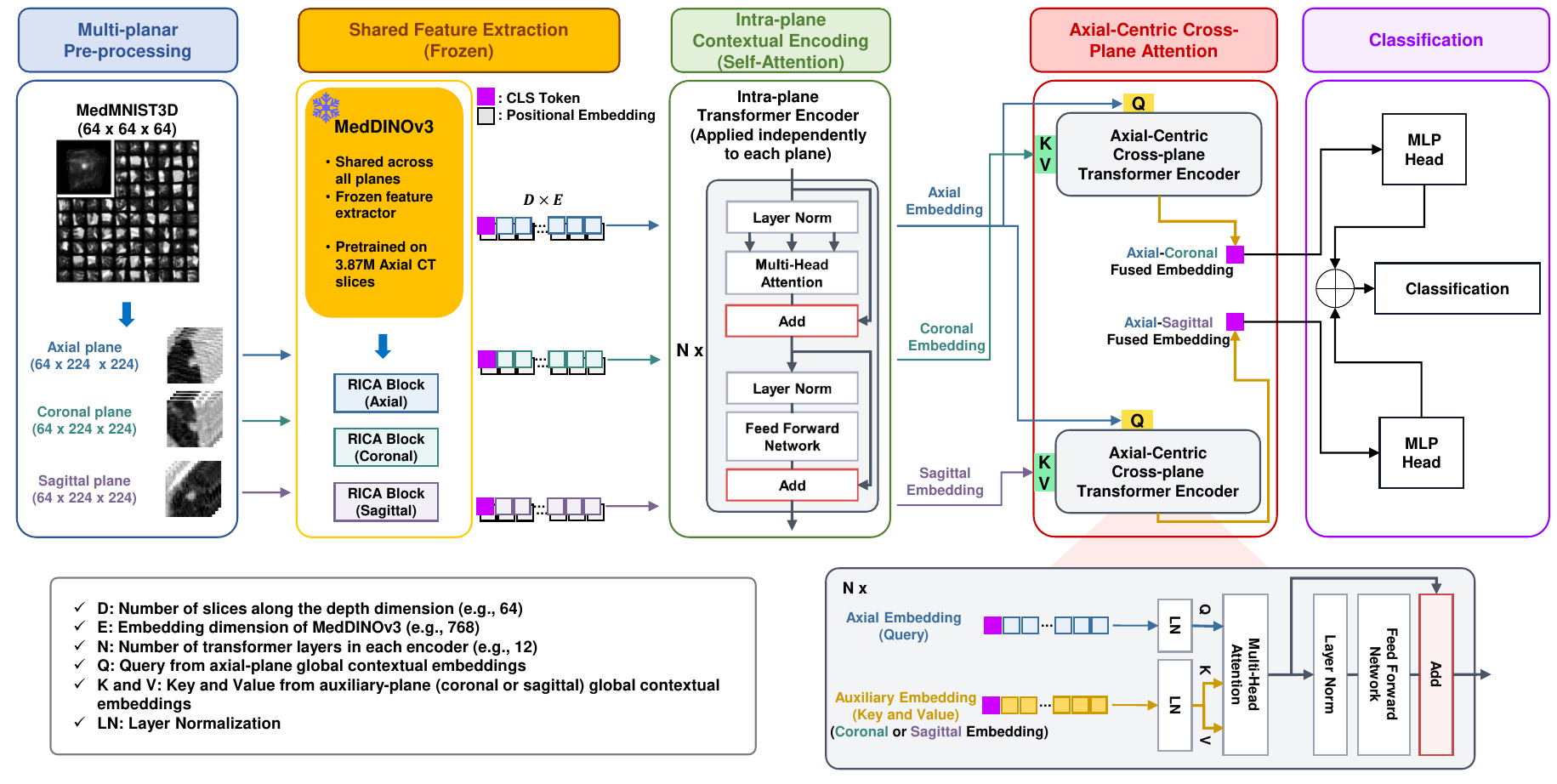}
   \caption{Overview of the proposed axial-centric cross-plane attention architecture. Multi-planar inputs pass through a frozen MedDINOv3 backbone for plane-wise feature extraction, followed by RICA blocks for local positional context modeling. Intra-plane transformer encoders independently capture contextual self-attention within each feature sequence. Axial-centric cross-plane transformer encoders condition the primary axial features (queries) on complementary coronal and sagittal information (keys and values). The fused axial-centric representations are passed to two multilayer perceptron (MLP) heads for final classification.}
\label{fig1}
\end{figure}

Several deep learning (DL) studies have demonstrated that multi-planar approaches outperform single-plane methods by leveraging complementary information across planes~\cite{num3,num4,num5,num6}. Similarly, multiple clinical studies have shown that supplementing the axial plane with coronal and sagittal planes improves diagnostic accuracy and clinician confidence~\cite{num7,num8,num9}. However, many existing multi-planar and 3D DL models treat all planes with equal importance or process the entire 3D volume, resulting in high computational costs and a limited ability to capture asymmetric dependencies between anatomical planes~\cite{num10,num11}. Consequently, these approaches often fail to reflect how clinicians selectively integrate information across planes, thus limiting the robustness and transferability of learned representations in current 3D and multi-planar methods.

Self-supervised vision foundation models (VFMs) pretrained on large-scale unlabeled data have demonstrated strong feature extractor capabilities for medical imaging tasks, despite being primarily pretrained on natural images~\cite{num12,num13}. However, VFMs pretrained on natural images exhibit a distinct feature representation gap compared to those VFMs pretrained on medical images, which limits feature transferability~\cite{num14}. To address these limitations, Li \etal. introduced MedDINOv3~\cite{num15}, a medical VFM pretrained on 3.87 million axial CT images using DINOv3, demonstrating strong performance in medical image segmentation.

Inspired by clinical interpretation workflows and the robust feature extraction capabilities of VFMs, we propose an axial-centric cross-plane attention architecture for 3D medical image classification. Our approach explicitly models asymmetric inter-plane dependencies between anatomical planes by selectively integrating anatomically relevant information extracted by a medical VFM from auxiliary planes into the axial plane via axial-centric cross-attention, aligning feature representations with clinically meaningful interpretation methods. Experiments on six datasets from the MedMNIST3D benchmark~\cite{num16} demonstrate that our proposed architecture achieves superior 3D classification performance compared to other architectures evaluated on the same datasets. Our key contributions are as follows:
\begin{enumerate}
    \item We introduce an axial-centric cross-plane attention architecture that employs a frozen medical VFM as the feature extractor across the axial, coronal, and sagittal planes, enabling robust 3D medical image classification under limited-data settings.
    \item The axial-centric cross-plane attention is an asymmetric inter-plane cross-attention mechanism that conditions axial feature representations on complementary information from auxiliary planes.
    \item Extensive experiments on diverse public 3D medical datasets demonstrate the superior performance of the proposed architecture. 
\end{enumerate}

\section{Methodology}
\subsection{Overall Architecture Overview}
As illustrated in Fig.~\ref{fig1}, the proposed architecture reflects an axial-centric clinical interpretation workflow and comprises a MedDINOv3 backbone, RICA blocks, intra-plane transformer encoders, axial-centric cross-plane transformer encoders, and multilayer perceptron (MLP) heads. The MedDINOv3 backbone extracts plane-specific 2D features from volumetric inputs across the axial, coronal, and sagittal planes.

To preserve positional context within each plane, the extracted features are processed by RICA blocks, which serve as CNN-based positional attention modules. Subsequently, transformer encoders are then applied independently to capture long-range contextual dependencies within each plane.

Following intra-plane self-attention, the axial features are independently conditioned on the coronal and sagittal features through axial-centric cross-plane transformer encoders. This asymmetric fusion selectively integrates complementary information from the auxiliary planes while maintaining the axial plane as the primary reference, consistent with clinical interpretation workflows. Finally, the classification (CLS) tokens from the fused axial-coronal and axial-sagittal representations are processed by MLP heads, and their logits are averaged to produce the final prediction.

\subsection{VFM-based Feature Extraction and Positional Context Modeling (RICA)}
A medical VFM, MedDINOv3, pretrained on large-scale axial CT images, is employed as a frozen feature extractor across all three anatomical planes as shown in Fig.~\ref{fig1}. The inputs consist of single-channel grayscale volumetric data with dimensions $B\times1\times D\times H\times W$, where B, D, H, and W represent the batch size, depth, height, and width of the input volume, respectively. Each volume is resized to $B\times3\times D\times224\times224$ by duplicating the grayscale channel to match MedDINOv3’s input resolution.

Each 2D slice is independently processed by MedDINOv3 to extract dense feature embeddings~\cite{num15}. The resulting slice-wise embeddings are then stacked along the depth dimension to form a feature sequence of size $B\times D\times E$, where E denotes the embedding dimension of MedDINOv3. While MedDINOv3 provides robust semantic feature representations at the slice level, highlighting informative positional patterns within the stacked feature sequence is essential for volumetric understanding. RICA blocks~\cite{num17} are applied to the stacked feature sequences to emphasize positional context by interpreting the $B\times D\times E$ sequences as a pseudo-2D feature map. The RICA block has previously demonstrated effectiveness in highlighting positional information and improving generalization to unseen data in 2D medical image segmentation tasks~\cite{num17,num18}. These steps are applied independently to the axial, coronal, and sagittal planes, producing plane-specific positional feature representations. Unless otherwise specified, $D = 64$ and $E = 768$ are used in all experiments.

\subsection{Intra-plane Transformer Encoder for Contextual Self-Attention}
To capture long-range contextual dependencies across non-adjacent slices within each anatomical plane, transformer encoders are applied independently to plane-specific stacked feature sequences. As illustrated in Fig.~\ref{fig1}, for each plane, the stacked feature sequence of size $B\times D\times E$ is used directly as the transformer input. Since the features are already embedded by MedDINOv3, no additional linear projection is applied. A learnable CLS token is prepended to aggregate global contextual information across slices. Learnable one-dimensional positional embeddings, including the CLS token, are added to provide spatial order information. The feature sequence from each plane is processed by a plane-specific transformer encoder consisting of $N = 12$ layers and $H = 12$ attention heads, following the standard Vision Transformer (ViT) architecture~\cite{num19}. This process is applied independently to the axial, coronal, and sagittal feature sequences. This design preserves plane-specific semantic feature representations while preventing premature fusion across anatomically asymmetric planes, resulting in three plane-specific global contextual embeddings.

\subsection{Axial-Centric Cross-Plane Transformer Encoder}
In clinical practice, lesion assessment is primarily conducted on the axial plane, with the coronal and sagittal planes subsequently referenced to provide complementary spatial context. Motivated by this asymmetric diagnostic workflow, we introduce an axial-centric cross-plane attention mechanism that conditions axial representations on information from the auxiliary planes while maintaining the axial plane as the primary reference.

Two axial-centric cross-plane transformer encoders are employed for independent fusion of axial features with coronal and sagittal features as shown in Fig.~\ref{fig1}. Each encoder processes the plane-specific global contextual embeddings produced by the intra-plane transformer encoders, which include a prepended CLS token. In each case, the axial embedding serves as the query, while the auxiliary-plane embedding (coronal or sagittal) serves as the keys and values, facilitating directional and selective fusion of complementary information rather than symmetric multi-plane aggregation. Each cross-plane transformer encoder is adapted from the intra-plane transformer encoder by replacing self-attention with cross-attention to enable inter-plane feature fusion. The residual connection following the cross-attention is intentionally omitted to reflect the asymmetric setting, where axial queries and auxiliary key-value pairs originate from semantically distinct representations. This omission prevents bias toward the original axial features and facilitates effective cross-plane conditioning. To maintain architectural consistency, the same number of encoder layers ($N = 12$) and attention heads ($H = 12$) as in the intra-plane transformer encoders are used. The CLS token is updated jointly with the slice tokens to aggregate axial-centric global contextual information conditioned on auxiliary planes. Through successive cross-attention layers, the axial query representations are iteratively refined using fixed auxiliary key–value representations, producing fused axial-coronal and axial-sagittal features. This dual cross-attention design facilitates axial-centric yet cross-plane contextual representations, reflecting the stepwise integration of auxiliary-plane information in clinical interpretation workflows.

Finally, the CLS tokens from the fused axial-coronal and axial-sagittal feature sequences are passed to two separate MLP heads. Each head consists of layer normalization followed by two fully connected layers with an embedding dimension $E$ and a $tanh$ activation function. The $tanh$ activation function is employed to improve training stability and mitigate overfitting, as MedMNIST3D is a lightweight 3D benchmark dataset. The resulting logits are then averaged to produce the final prediction.

\section{Experiments}
\subsection{Dataset and Experimental Setup}
We evaluate the proposed axial-centric cross-plane attention architecture using the MedMNIST+ benchmark, an expanded version of MedMNISTv2~\cite{num16,num26}. Within MedMNIST+, MedMNIST3D serves as a standardized benchmark for 3D medical image classification. The datasets cover diverse medical imaging modalities, including CT, magnetic resonance angiography (MRA), and electron microscopy, supporting both binary and multi-class classification tasks. Each dataset is preprocessed into a consistent format and provided with predefined training, validation, and test splits to ensure fair comparisons across methods. In this study, we evaluate our architecture on all six MedMNIST3D datasets, each with volumetric dimensions of $64\times64\times64$. OrganMNIST3D consists of 1,742 abdominal CT volumes with 11 classes (971/161/610 for training/validation/test). FractureMNIST3D comprises 1,370 chest CT volumes with three classes (1,027/103/240). VesselMNIST3D includes 1,908 shapes from brain MRA volumes with two classes (1,335/191/382). SynapseMNIST3D consists of 1,759 electron microscope volumes with two classes (1,230/177/352). NoduleMNIST3D includes 1,633 chest CT volumes with two classes (1,158/165/310). AdrenalMNIST3D contains 1,584 shapes from abdominal CT volumes with two classes (1,188/98/298). 

PyTorch container (2.3.0a0+40ec155e58.nv24.03) with Python 3.10.12 was used for all experiments. The experiments were conducted on an NVIDIA DGX-A100 GPU (40 GB) and four CPU cores (64 GB) of an AMD EPYC 7742 processor. The proposed architecture was trained for 100 epochs with a batch size of 4. The Adam optimizer with a cosine annealing warm restart scheduler was employed with an initial learning rate of 1e-12 and a warm-up period of 5 epochs. The maximum learning rate was set to 1e-4 for NoduleMNIST3D and 1e-5 for the remaining datasets. Cross-entropy loss was used for multi-class classification, while binary cross-entropy with logits loss was applied for binary tasks. To improve generalization, six data augmentation methods from the TorchIO library were selectively applied during training: RandomAnisotropy, RandomAffine, RandomFlip, RandomNoise, RandomBlur, and RandomGamma.

\begin{table}[t]
\begin{center}
\scalebox{0.6}{
\begin{tabular}{c|ccccccccccccc}
\hline
 &
   \# of trainable&
  \multicolumn{2}{c}{Fracture3D} &
  \multicolumn{2}{c}{Adrenal3D} &
  \multicolumn{2}{c}{Nodule3D} &
  \multicolumn{2}{c}{Vessel3D} &
  \multicolumn{2}{c}{Organ3D} &
  \multicolumn{2}{c}{Synapse3D} \\ \cline{3-14}
Methods                                                        & params & ACC  & AUC  & ACC  & AUC  & ACC  & AUC  & ACC  & AUC  & ACC  & AUC  & ACC  & AUC  \\ \hline
Official~\cite{num16}             & 33M    & 50.8 & 71.2 & 72.1 & 82.7 & 84.4 & 86.3 & 87.7 & 87.4 & 90.7 & 99.6 & 74.5 & 82.0 \\
BSDA~\cite{num20}                 & 33M    & 56.9 & \textbf{73.1} & 83.8 & 89.2 & 86.1 & 89.2 & 93.2 & 91.7 & 88.7 & 99.4 & -    & -    \\
ViViT-M + R-LLM~\cite{num21}      & 295M    & 56.3 & 66.9 & 83.2 & 84.9 & 87.4 & 88.5 & 90.6 & 87.0 & -    & -    & -    & -    \\
LMTTM-VMI~\cite{num22}            & 21M   & 56.2 & 69.1 & 84.2 & 84.7 & 88.3 & 89.1 & 93.1 & 91.7 & 94.8 & 99.7 & 77.7 & 73.7 \\
Medformer~\cite{num23}            & -   & 42.7 & 53.6 & -    & -    & 84.2 & 87.7 & -    & -    & 93.7 & 99.1 & -    & -    \\
CdTransformer + CcCL~\cite{num24} & -  & 52.9 & 72.4 & 83.6 & 88.4 & \textbf{90.3} & \textbf{94.3} & 92.9 & \textbf{95.9} & -    & -    & 83.2 & 87.9 \\ \hline
AC-Tiny  & 9.5M  & 55.0 & 68.6 & 87.6 & 91.2 & 88.1 & 90.4 & \textbf{95.5} & 92.8 & 95.7 & 99.9 & 83.2 & 84.6 \\
AC-Small & 216M  & 56.3 & 71.7 & 87.2 & 91.8 & 87.4 & 82.1 & 94.5 & 90.9 & 95.2 & 99.8 & 84.1 & 87.5 \\
AC-Base  & 427M  & \textbf{60.8} & 71.9 & \textbf{88.6} & \textbf{92.0} & 89.0 & 90.2 & 95.0 & 93.3 & \textbf{95.7} & \textbf{99.9} & \textbf{86.1} & \textbf{88.0}    \\ \hline
\end{tabular}}
\end{center}
\caption{Quantitative comparison of the proposed axial-centric cross-plane attention model variants with six existing 3D medical image classification models on the MedMNIST3D benchmark. The proposed variants include Axial-Centric-Tiny (AC-Tiny, $N = 4$, $H = 8$, $E = 192$), Axial-Centric-Small (AC-Small, $N = 6$, $H = 4$, $E = 768$), and Axial-Centric-Base (AC-Base, $N = 12$, $H = 12$, $E = 768$). The best results are shown in \textbf{bold}. For conciseness, the suffix MNIST is omitted from dataset names.}
\label{tab1}
\end{table}

\subsection{Comparison with Other Methods}
We evaluated the proposed architecture on the MedMNIST3D benchmark and compared its performance against six existing methods: a ResNet-18 backbone reported by the official MedMNISTv2 authors~\cite{num16}, BSDA~\cite{num20}, ViViT-M + R-LLM~\cite{num21}, LMTTM-VMI~\cite{num22}, Medformer~\cite{num23}, and CdTransformer + CcCL~\cite{num24}. To represent model capacity, we report the number of trainable parameters for our models alongside accuracy (ACC) and area under the ROC curve (AUC) as evaluation metrics. For multi-class classification tasks, we report macro-averaged one-vs-rest AUC. The quantitative results are presented in Table~\ref{tab1}.

The proposed architecture was further evaluated using three model variants: Axial-Centric-Tiny (AC-Tiny, $N = 4$, $H = 8$, $E = 192$), Axial-Centric-Small (AC-Small, $N = 6$, $H = 4$, $E = 768$), and Axial-Centric-Base (AC-Base, $N = 12$, $H = 12$, $E = 768$). AC-Tiny contains 9.5 million trainable parameters and 86 million frozen MedDINOv3 parameters; AC-Small contains 216 million trainable parameters and 86 million frozen MedDINOv3 parameters; and AC-Base contains 427 million trainable parameters and 86 million frozen MedDINOv3 parameters. AC-Tiny employs an additional linear projection layer to reduce the embedding dimension of MedDINOv3 from 768 to 192 before intra-plane transformer encoders. Compared to ResNet-18/BDSA (33 million), LMTTM-VMI (21 million), and ViViT-M + R-LLM (295 million), the proposed models cover a wide range of model capacities while maintaining the same frozen feature extractor. Despite having substantially fewer trainable parameters, AC-Tiny remained highly competitive, achieving the highest accuracy of 95.5\% on VesselMNIST3D. AC-Base achieved the highest accuracy on four MedMNIST3D datasets and the highest AUC on three MedMNIST3D datasets. Notably, it outperformed the second-best method by 3.9\% in accuracy on FractureMNIST3D. On NoduleMNIST3D, AC-Base achieved the second-highest accuracy and AUC, trailing the best-performing method by 1.3\% and 4.1\%, respectively. These results indicate that the axial-centric cross-plane attention design effectively captures complementary information across planes, even with reduced trainable capacity, while the larger variant further benefits from increased representational power across diverse modalities, including CT, brain MRA, and electron microscopy.

\subsection{Interpretability Analysis}
To qualitatively assess how the proposed architecture identifies anatomically relevant information across different planes, we conducted a slice-level feature attribution analysis on NoduleMNIST3D, VesselMNIST3D, and FractureMNIST3D datasets using Grad-CAM~\cite{num27}-based visualizations derived from the frozen MedDINOv3 backbone. Since the proposed model processes stacked feature sequences rather than image-space inputs after feature extraction, the interpretability analysis was performed using the plane-specific intra-plane transformer representations prior to cross-plane fusion.

\begin{figure}
\begin{center}
\end{center}
\includegraphics[width=\textwidth]{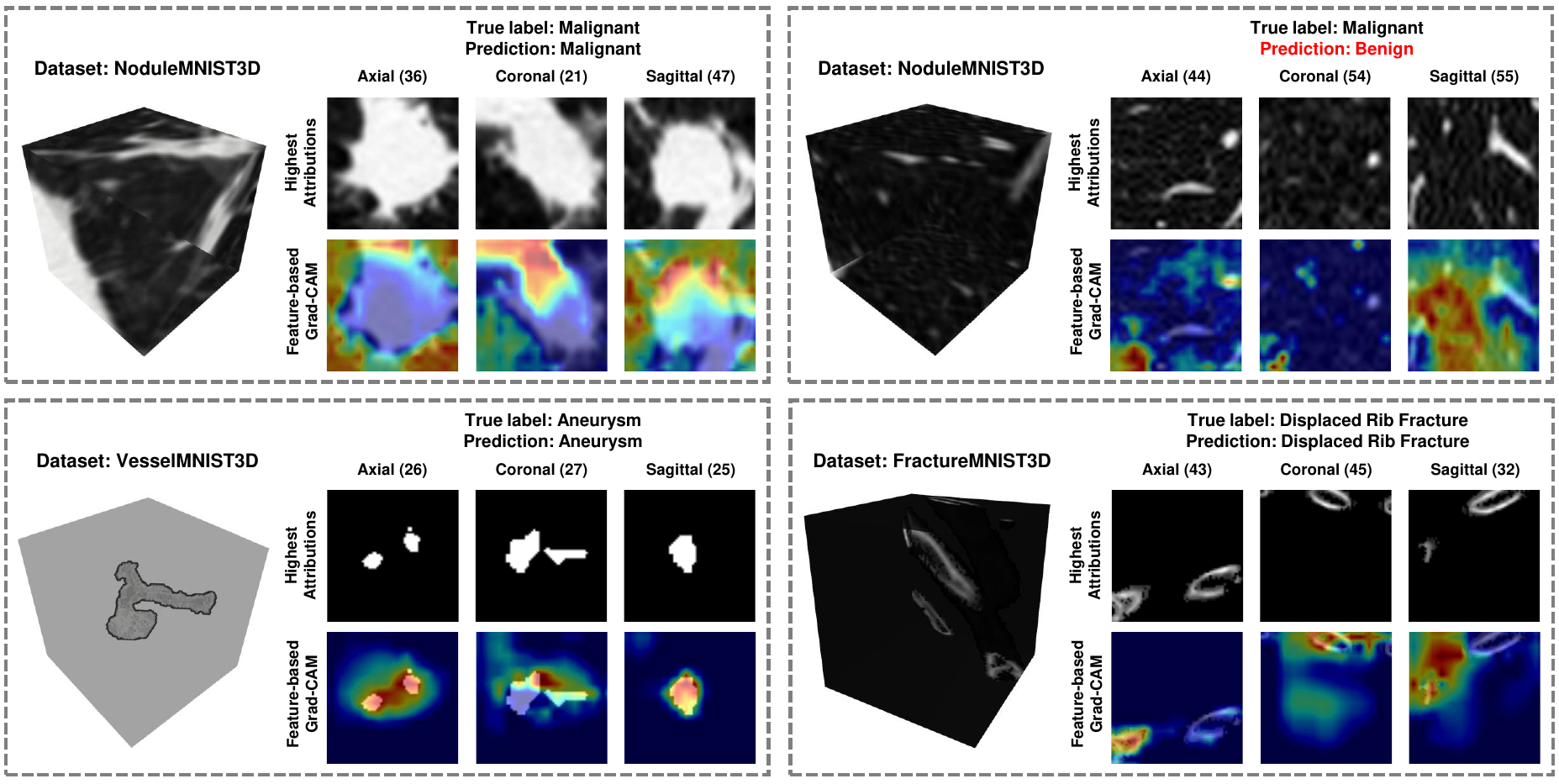}
   \caption{Slice-level Grad-CAM heatmaps across three MedMNIST3D benchmarks. The slices correspond to the highest attributions from each intra-plane transformer encoder, with the selected indices indicated in parentheses (warmer colors represent stronger attribution). Correctly classified cases exhibit anatomically coherent localization across all three planes, validating effective slice-level selection before cross-plane fusion. In contrast, the misclassified malignant lung nodule shows comparatively scattered and poorly localized heatmaps. Notably, although MedDINOv3 was exclusively pretrained on axial CT slices, anatomically coherent localization is consistently observed across the coronal and sagittal planes, suggesting that feature representations are transferable across auxiliary planes.}
\label{fig2}
\end{figure}

MedDINOv3 first extracts slice-wise feature embeddings from the axial, coronal, and sagittal planes, producing stacked feature sequences of size $D\times E$. These plane-specific sequences are then processed by intra-plane transformer encoders, where each token corresponds to an individual anatomical slice. To identify the most informative slice within each plane, slice-level importance scores were derived from the intra-plane transformer encoder representations using gradient-based attribution. The slice with the highest attribution from each anatomical plane was then passed through the frozen MedDINOv3 backbone to generate feature-based Grad-CAM heatmaps from the backbone embedding representations. Since MedDINOv3 remained frozen during training, these heatmaps reflect the pretrained backbone representations rather than task-adapted feature representations learned during downstream training. Therefore, this analysis primarily evaluates whether the intra-plane transformer encoders successfully identify slices containing diagnostically relevant anatomical information. Interpretability analysis was intentionally conducted before the cross-plane transformer encoders because cross-plane attention operates on fused latent representations across anatomical planes. As a result, the fused tokens no longer maintain a one-to-one correspondence with individual slices along each anatomical axis, rendering slice-level attribution ambiguous.

Figure~\ref{fig2} presents Grad-CAM heatmaps from the axial, coronal, and sagittal planes. In correctly classified cases, the selected slices consistently localize anatomically relevant regions, demonstrating that the intra-plane transformer encoders effectively identify informative slice-level representations before cross-plane fusion. Notably, although MedDINOv3 was exclusively pretrained on large-scale axial CT slices, the coronal and sagittal attribution maps also exhibit coherent anatomical localization. This suggests that the pretrained backbone retains transferable feature representations across auxiliary planes. 

In contrast, the misclassified malignant lung nodule case exhibits relatively scattered attributions throughout the background rather than on the lesion itself, which may have contributed to the incorrect benign prediction. Overall, these findings indicate that the proposed architecture preserves anatomically meaningful feature localization while selectively integrating plane-specific contextual information. However, misclassifications may result from less discriminative attribution patterns.

\begin{figure}
\begin{center}
\end{center}
\includegraphics[width=\textwidth]{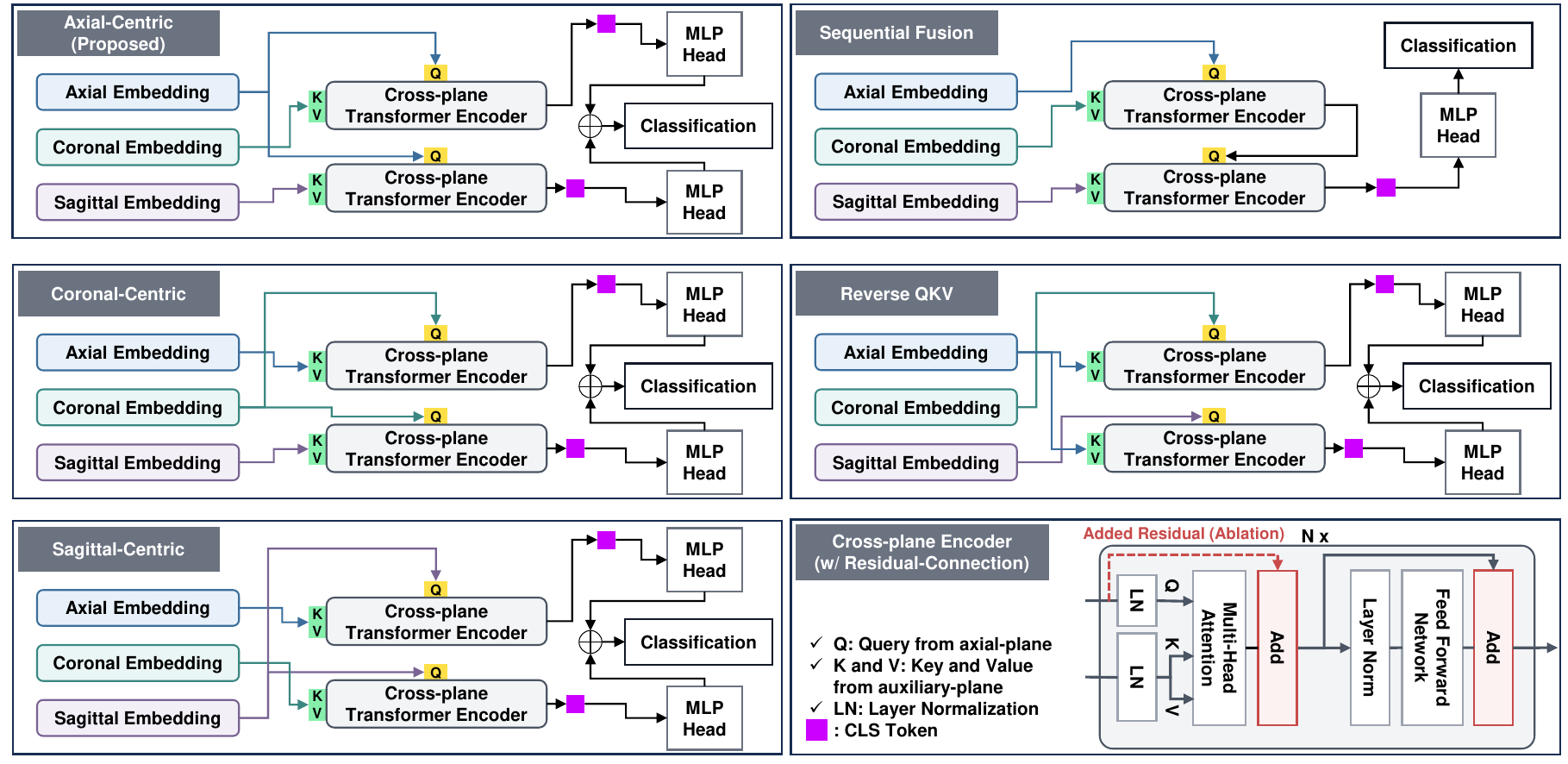}
   \caption{Comparison of ablation study designs for axial-centric fusion and architectural modifications. The figure illustrates the Sequential Fusion, Coronal-Centric, Sagittal-Centric, and Reverse query-key-value (QKV) variants, as well as the residual connection in cross-attention design. The proposed architecture employs axial-centric querying and intentionally omits the residual connection following cross-attention, whereas the residual cross-attention variant incorporates the residual pathway, highlighted by the red dotted box.}
\label{fig3}
\end{figure}

\subsection{Axial-Centric Fusion Design Ablations}
To validate the proposed axial-centric fusion strategy, we conducted ablation studies comparing alternative cross-plane fusion designs, including sequential fusion and coronal- and sagittal-centric querying strategies. The corresponding fusion designs are illustrated in Fig.~\ref{fig3}, and the quantitative results are summarized in Table~\ref{tab2}. Replacing the proposed dual axial-centric cross-attention modules with a two-stage sequential axial–coronal–sagittal fusion resulted in modest performance decreases across most datasets, although marginal improvements in AUC were observed on FractureMNIST3D and VesselMNIST3D. This suggests that independently conditioning axial features on each auxiliary plane is more effective than sequential multi-plane conditioning. Additionally, coronal- and sagittal-centric querying both led to further performance declines across most datasets, indicating that the axial plane is the most effective reference for directional cross-plane conditioning in this context. Overall, these results highlight the importance of axial-centric and directional cross-plane fusion.

\begin{table}[t]
\begin{center}
\scalebox{0.7}{
\begin{tabular}{c|cccccccccccc}
\hline
& 
\multicolumn{2}{c}{Fracture3D} & \multicolumn{2}{c}{Adrenal3D} & \multicolumn{2}{c}{Nodule3D} & \multicolumn{2}{c}{Vessel3D} & \multicolumn{2}{c}{Organ3D} & \multicolumn{2}{c}{Synapse3D} \\ \cline{2-13}
Methods & ACC & AUC & ACC & AUC & ACC & AUC & ACC & AUC & ACC & AUC & ACC & AUC \\ \hline
Sequential Fusion& 60.4 & \textbf{73.9} & 88.3 & 91.6 & 86.8 & 84.8 & 94.0 & 93.7 & 94.4 & 99.8 & 83.5 & 84.8 \\
Coronal-Centric& 57.5 & 72.3 & 87.6 & 88.6 & 87.4 & 87.6 & 94.8 & 93.9 & 95.1 & 99.8 & 84.1 & 85.9 \\ 
Sagittal-Centric& 57.5 & 72.6 & 86.2 & 86.7 & 86.1 & 78.3 & 94.5 & \textbf{95.6} & 94.6 & 99.7 & 84.1 & 87.6 \\ \hline
AC-Base & \textbf{60.8} & 71.9 & \textbf{88.6} & \textbf{92.0} & \textbf{89.0} & \textbf{90.2} & \textbf{95.0} & 93.3 & \textbf{95.7} & \textbf{99.9} & \textbf{86.1} & \textbf{88.0} \\ \hline
\end{tabular}}
\end{center}
\caption{Ablation studies evaluating the axial-centric fusion design on the MedMNIST3D benchmark. All variants employ $N = H = 12$ and $E = 768$. The best results are shown in \textbf{bold}. For conciseness, the suffix MNIST is omitted from dataset names. Figure~\ref{fig3} illustrates the sequential, coronal-centric, and sagittal-centric fusion variants. The sequential variant performs a two-stage axial–coronal–sagittal fusion, while the coronal-centric and sagittal-centric variants use the corresponding plane as the query. The results highlight the importance of axial-centric and directional cross-plane conditioning. }
\label{tab2} 
\end{table}

\subsection{Architectural Design Ablations}
To further analyze the impact of architectural design choices, we conducted ablation studies on residual connections in cross-attention, activation functions in the classification head, and query–key–value (QKV) allocation. The quantitative results are summarized in Table~\ref{tab3}. Introducing a residual connection after cross-attention consistently degraded performance across most datasets, indicating that omitting this residual connection better supports asymmetric cross-plane conditioning by reducing bias toward the original axial features. Replacing the $tanh$ activation function in the classification head with GeLU generally reduced performance, suggesting that $tanh$ is better suited for the lightweight classification heads used in this context. Finally, reversing the proposed QKV allocation, using auxiliary-plane features as queries and axial features as keys and values as shown in Fig.~\ref{fig3}, consistently resulted in substantial performance declines across all datasets. This included a 6.2\% accuracy drop on OrganMNIST3D and a 5.9\% AUC reduction on AdrenalMNIST3D. These results highlight the importance of employing the axial feature sequence as the query to enable effective axial-centric conditioning. 

Overall, these results support that both the cross-attention design and the classification-head activation contribute to the proposed architecture, with axial-centric QKV allocation introducing the strongest effect.

\begin{table}[t]
\begin{center}
\scalebox{0.65}{
\begin{tabular}{c|cccccccccccc}
\hline
& 
\multicolumn{2}{c}{Fracture3D} & \multicolumn{2}{c}{Adrenal3D} & \multicolumn{2}{c}{Nodule3D} & \multicolumn{2}{c}{Vessel3D} & \multicolumn{2}{c}{Organ3D} & \multicolumn{2}{c}{Synapse3D} \\ \cline{2-13}
Methods & ACC & AUC & ACC & AUC & ACC & AUC & ACC & AUC & ACC & AUC & ACC & AUC \\ \hline
Cross-Attn + Residual & 57.1 & 66.4 & 87.6 & 91.9 & 85.2 & 87.7 & 94.8 & \textbf{93.7} & 95.6 & 99.9 & 84.9 & 86.6 \\
GeLU Head & 56.7 & 68.4 & 86.9 & 89.9 & 87.1 & 86.0 & 94.8 & 91.0 & 95.2 & 99.8 & 83.5 & 86.7 \\
Reverse QKV & 57.1 & 69.9 & 84.9 & 86.1 & 86.5 & 87.1 & 94.2 & 93.2 & 89.5 & 99.3 & 83.0 & 84.9 \\ \hline
AC-Base & \textbf{60.8} & \textbf{71.9} & \textbf{88.6} & \textbf{92.0} & \textbf{89.0} & \textbf{90.2} & \textbf{95.0} & 93.3 & \textbf{95.7} & \textbf{99.9} & \textbf{86.1} & \textbf{88.0} \\ \hline
\end{tabular}}
\end{center}
\caption{Ablation studies evaluating the residual connection in cross-attention, the activation function of the classification head, and query-key-value (QKV) allocation on the MedMNIST3D benchmark. All variants employ $N = H = 12$ and $E = 768$. The best results are shown in \textbf{bold}. For conciseness, the suffix MNIST is omitted from dataset names. The results highlight the importance of omitting the residual connection after cross-attention, employing axial-centric QKV allocation, and using the $tanh$ activation function in the classification head.}
\label{tab3} 
\end{table}

\section{Conclusion, Limitations, and Future Work}
In this work, we proposed an axial-centric cross-plane attention architecture for 3D medical image classification, inspired by axial-primary clinical interpretation workflows. The proposed architecture integrates frozen MedDINOv3 feature extraction with intra-plane contextual self-attention and directional cross-plane attention to selectively condition axial feature representations on complementary coronal and sagittal information. By explicitly modeling asymmetric inter-plane dependencies, the architecture learns axial-centric yet cross-plane contextual representations that align with a multi-planar clinical interpretation workflow. Extensive experiments on the MedMNIST3D benchmark demonstrate that the proposed architecture consistently outperforms existing 3D and multi-plane classification methods across diverse imaging modalities. Furthermore, the introduction of a lightweight AC-Tiny variant demonstrates that these significant performance improvements are driven by our clinically aligned architectural design, providing a computationally efficient alternative for resource-constrained environments. Additional ablation studies further confirm the importance of axial-centric querying and QKV allocation, directional cross-plane fusion, the omission of residual connections in cross-attention, and specific architectural design choices in achieving robust performance improvements. Interpretability analysis also reveals that the proposed intra-plane contextual attention mechanism identifies diagnostically relevant slice-level representations prior to cross-plane fusion. Overall, these findings highlight the importance of aligning architectural design with clinical interpretation patterns and suggest that axial-centric multi-planar representation learning is an effective framework representation learning for volumetric medical image classification.

Although the proposed architecture demonstrates strong performance on the MedMNIST3D benchmark, this study has several limitations. First, while our primary base architecture (AC-Base) achieves the highest diagnostic accuracies across four datasets, it introduces a substantial number of trainable parameters due to the multiple intra-plane and cross-plane transformer encoders. Although the AC-Tiny variant demonstrates competitive performance with only 9.5 million trainable parameters, balancing the trade-off between absolute performance and deployment in resource-constrained clinical environments remains a challenge for real-time clinical application. Second, the frozen MedDINOv3 backbone was pretrained exclusively on axial CT slices, potentially introducing an axial-representation bias when applied to coronal and sagittal planes. Although interpretability analysis suggests transferable feature localization across auxiliary planes, the potential impact of axial-pretraining bias on cross-plane representation learning remains insufficiently explored. Third, all volumetric inputs were resized from $64\times64$ to $224\times224$ to match the MedDINOv3 input resolution. This resizing may introduce interpolation artifacts and alter fine-grained anatomical structures, particularly in small lesions or low-resolution datasets. Additionally, the proposed axial-centric design was primarily motivated by CT interpretation workflows and may not generalize equally well to modalities or anatomical regions where coronal or sagittal planes play a more dominant diagnostic role.

In future work, we plan to evaluate the proposed architecture on higher-resolution institutional cardiac CT and MRI datasets, where multi-planar interpretation plays a central diagnostic role. Additionally, we aim to investigate alternative large-scale, medical data pretrained models, such as RadImageNet~\cite{num25}, as well as coronal and sagittal CT pretrained models, to serve as feature extractors within the proposed architecture. This will allow us to further examine potential axial-pretraining bias and improve cross-plane representation learning in volumetric medical imaging.

\section{Acknowledgments}
The computational work was fully supported and performed on resources of the National Supercomputing Centre (NSCC), Singapore (https://www.nscc.sg), and CHROMA in SingHealth Tower, SingHealth, Singapore (https://www.singhealth.com.sg).

\bibliography{egbib}

\begin{thebibliography}{27}
\providecommand{\natexlab}[1]{#1}
\providecommand{\url}[1]{\texttt{#1}}
\expandafter\ifx\csname urlstyle\endcsname\relax
  \providecommand{\doi}[1]{doi: #1}\else
  \providecommand{\doi}{doi: \begingroup \urlstyle{rm}\Url}\fi

\bibitem[Abdelhamid et~al.(2024)Abdelhamid, Bhatt, Viana, Ferreira, Nogueira, Al-Bayati, Grossberg, Allen, and Haussen]{num8}
Hend~M Abdelhamid, Nirav~R Bhatt, Lorena~S Viana, Felipe~M Ferreira, Raul~G Nogueira, Alhamza~R Al-Bayati, Jonathan~A Grossberg, Jason~W Allen, and Diogo~C Haussen.
\newblock Multiplane reconstruction modifies the diagnostic performance of ct angiography in carotid webs.
\newblock \emph{Clinical neurology and neurosurgery}, 244:\penalty0 108441, 2024.

\bibitem[Angkoso et~al.(2022)Angkoso, Tjahyaningtijas, Adrianto, Sensusiati, Purnama, and Purnomo]{num11}
Cucun~Very Angkoso, Hapsari Peni~Agustin Tjahyaningtijas, Yudhi Adrianto, Anggraini~Dwi Sensusiati, I~Ketut~Eddy Purnama, and Mauridhi~Hery Purnomo.
\newblock Multi-features fusion in multi-plane mri images for alzheimer’s disease classification.
\newblock \emph{Int. J. Intell. Eng. Syst}, 15\penalty0 (4):\penalty0 182--197, 2022.

\bibitem[Baharoon et~al.(2023)Baharoon, Qureshi, Ouyang, Xu, Aljouie, and Peng]{num14}
Mohammed Baharoon, Waseem Qureshi, Jiahong Ouyang, Yanwu Xu, Abdulrhman Aljouie, and Wei Peng.
\newblock Evaluating general purpose vision foundation models for medical image analysis: An experimental study of dinov2 on radiology benchmarks.
\newblock \emph{arXiv preprint arXiv:2312.02366}, 2023.

\bibitem[Black and Souvenir(2024)]{num5}
Samuel Black and Richard Souvenir.
\newblock Multi-view classification using hybrid fusion and mutual distillation.
\newblock In \emph{Proceedings of the IEEE/CVF Winter Conference on Applications of Computer Vision}, pages 270--280, 2024.

\bibitem[Dosovitskiy et~al.(2020)Dosovitskiy, Beyer, Kolesnikov, Weissenborn, Zhai, Unterthiner, Dehghani, Minderer, Heigold, Gelly, et~al.]{num19}
Alexey Dosovitskiy, Lucas Beyer, Alexander Kolesnikov, Dirk Weissenborn, Xiaohua Zhai, Thomas Unterthiner, Mostafa Dehghani, Matthias Minderer, Georg Heigold, Sylvain Gelly, et~al.
\newblock An image is worth 16x16 words: Transformers for image recognition at scale.
\newblock \emph{arXiv preprint arXiv:2010.11929}, 2020.

\bibitem[Jang and Hwang(2022)]{num10}
Jinseong Jang and Dosik Hwang.
\newblock M3t: three-dimensional medical image classifier using multi-plane and multi-slice transformer.
\newblock In \emph{Proceedings of the IEEE/CVF conference on computer vision and pattern recognition}, pages 20718--20729, 2022.

\bibitem[Lai et~al.(2024)Lai, Wu, Chen, Zhou, and Hovakimyan]{num21}
Zhixin Lai, Jing Wu, Suiyao Chen, Yucheng Zhou, and Naira Hovakimyan.
\newblock Residual-based language models are free boosters for biomedical imaging tasks.
\newblock In \emph{Proceedings of the IEEE/CVF Conference on Computer Vision and Pattern Recognition}, pages 5086--5096, 2024.

\bibitem[Li et~al.(2024)Li, Hu, and Yang]{num12}
Yuheng Li, Mingzhe Hu, and Xiaofeng Yang.
\newblock Polyp-sam: Transfer sam for polyp segmentation.
\newblock In \emph{Medical imaging 2024: computer-aided diagnosis}, volume 12927, pages 749--754. SPIE, 2024.

\bibitem[Li et~al.(2025)Li, Wu, Lai, Hu, and Yang]{num15}
Yuheng Li, Yizhou Wu, Yuxiang Lai, Mingzhe Hu, and Xiaofeng Yang.
\newblock Meddinov3: How to adapt vision foundation models for medical image segmentation?
\newblock \emph{arXiv preprint arXiv:2509.02379}, 2025.

\bibitem[Liu et~al.(2025)Liu, Chen, Shi, Lu, Jian, Pan, Cai, Wang, Yu, Gao, et~al.]{num13}
Che Liu, Yinda Chen, Haoyuan Shi, Jinpeng Lu, Bailiang Jian, Jiazhen Pan, Linghan Cai, Jiayi Wang, Jieming Yu, Ziqi Gao, et~al.
\newblock Does dinov3 set a new medical vision standard? benchmarking 2d and 3d classification, segmentation, and registration.
\newblock \emph{arXiv preprint arXiv:2509.06467}, 2025.

\bibitem[Mang et~al.(2009)Mang, Schaefer-Prokop, Schima, Maier, Schober, Mueller-Mang, Weber, and Prokop]{num2}
Thomas Mang, Cornelia Schaefer-Prokop, Wolfgang Schima, Andrea Maier, Ewald Schober, Christina Mueller-Mang, Michael Weber, and Mathias Prokop.
\newblock Comparison of axial, coronal, and primary 3d review in mdct colonography for the detection of small polyps: a phantom study.
\newblock \emph{European journal of radiology}, 70\penalty0 (1):\penalty0 86--93, 2009.

\bibitem[Mcmenamin et~al.(2015)Mcmenamin, Pearce, and Klassen]{num7}
Drew Mcmenamin, Alex Pearce, and Matthew Klassen.
\newblock Visual search in abdominopelvic ct interpretation: accuracy and time efficiency between coronal mpr and axial images.
\newblock \emph{Academic Radiology}, 22\penalty0 (2):\penalty0 164--168, 2015.

\bibitem[Mei et~al.(2022)Mei, Liu, Robson, Marinelli, Huang, Doshi, Jacobi, Cao, Link, Yang, et~al.]{num25}
Xueyan Mei, Zelong Liu, Philip~M Robson, Brett Marinelli, Mingqian Huang, Amish Doshi, Adam Jacobi, Chendi Cao, Katherine~E Link, Thomas Yang, et~al.
\newblock Radimagenet: an open radiologic deep learning research dataset for effective transfer learning.
\newblock \emph{Radiology: Artificial Intelligence}, 4\penalty0 (5):\penalty0 e210315, 2022.

\bibitem[Park et~al.(2025{\natexlab{a}})Park, Kim, Chang, Leng, Zhong, and Baskaran]{num17}
Doyoung Park, Jinsoo Kim, Qi~Chang, Shuang Leng, Liang Zhong, and Lohendran Baskaran.
\newblock Ricau-net: Residual-block inspired coordinate attention u-net for segmentation of small and sparse calcium lesions in cardiac ct.
\newblock In \emph{2025 IEEE 22nd International Symposium on Biomedical Imaging (ISBI)}, pages 1--5. IEEE, 2025{\natexlab{a}}.

\bibitem[Park et~al.(2025{\natexlab{b}})Park, Ng, Zhong, Tan, Wang, Lai, Zhong, Ooi, Tan, and Baskaran]{num18}
Doyoung Park, Jedidiah Ng, Yixin Zhong, Chun Sheng~Alvin Tan, Xiaomeng Wang, Gillianne Geet~Yi Lai, Liang Zhong, Su~Kai~Gideon Ooi, Daniel Shao~Weng Tan, and Lohendran Baskaran.
\newblock Evaluating the generalizability of an automated coronary artery calcium segmentation and scoring algorithm using multi-vendor dataset.
\newblock \emph{Scientific Reports}, 15\penalty0 (1):\penalty0 21744, 2025{\natexlab{b}}.

\bibitem[Sandrasegaran et~al.(2007)Sandrasegaran, Rydberg, Tann, Hawes, Kopecky, and Maglinte]{num9}
K~Sandrasegaran, J~Rydberg, M~Tann, DR~Hawes, KK~Kopecky, and DD~Maglinte.
\newblock Benefits of routine use of coronal and sagittal reformations in multi-slice ct examination of the abdomen and pelvis.
\newblock \emph{Clinical radiology}, 62\penalty0 (4):\penalty0 340--347, 2007.

\bibitem[Selvaraju et~al.(2017)Selvaraju, Cogswell, Das, Vedantam, Parikh, and Batra]{num27}
Ramprasaath~R Selvaraju, Michael Cogswell, Abhishek Das, Ramakrishna Vedantam, Devi Parikh, and Dhruv Batra.
\newblock Grad-cam: Visual explanations from deep networks via gradient-based localization.
\newblock In \emph{Proceedings of the IEEE international conference on computer vision}, pages 618--626, 2017.

\bibitem[Setio et~al.(2016)Setio, Ciompi, Litjens, Gerke, Jacobs, Van~Riel, Wille, Naqibullah, S{\'a}nchez, and Van~Ginneken]{num3}
Arnaud Arindra~Adiyoso Setio, Francesco Ciompi, Geert Litjens, Paul Gerke, Colin Jacobs, Sarah~J Van~Riel, Mathilde Marie~Winkler Wille, Matiullah Naqibullah, Clara~I S{\'a}nchez, and Bram Van~Ginneken.
\newblock Pulmonary nodule detection in ct images: false positive reduction using multi-view convolutional networks.
\newblock \emph{IEEE transactions on medical imaging}, 35\penalty0 (5):\penalty0 1160--1169, 2016.

\bibitem[Simionescu(2025)]{num23}
Cristian Simionescu.
\newblock Medformer: A multitask multimodal foundational model for medical imaging.
\newblock \emph{Procedia Computer Science}, 270:\penalty0 446--455, 2025.

\bibitem[Ter-Pogossian(1977)]{num1}
Michel~M Ter-Pogossian.
\newblock Basic principles of computed axial tomography.
\newblock In \emph{Seminars in nuclear medicine}, volume 7, number 2, pages 109--127. Elsevier, 1977.

\bibitem[Van~Tulder et~al.(2021)Van~Tulder, Tong, and Marchiori]{num4}
Gijs Van~Tulder, Yao Tong, and Elena Marchiori.
\newblock Multi-view analysis of unregistered medical images using cross-view transformers.
\newblock In \emph{International Conference on Medical Image Computing and Computer-Assisted Intervention}, pages 104--113. Springer, 2021.

\bibitem[Wei et~al.(2025)Wei, Yang, Sun, Feng, Wang, and Han]{num22}
Hongkai Wei, Yang Yang, Shijie Sun, Mingtao Feng, Rong Wang, and Xianfeng Han.
\newblock Lmttm-vmi: Linked memory token turing machine for 3d volumetric medical image classification.
\newblock \emph{Computer Methods and Programs in Biomedicine}, 262:\penalty0 108640, 2025.

\bibitem[Yang et~al.(2021)Yang, Shi, and Ni]{num26}
Jiancheng Yang, Rui Shi, and Bingbing Ni.
\newblock Medmnist classification decathlon: A lightweight automl benchmark for medical image analysis.
\newblock In \emph{2021 IEEE 18th international symposium on biomedical imaging (ISBI)}, pages 191--195. IEEE, 2021.

\bibitem[Yang et~al.(2023)Yang, Shi, Wei, Liu, Zhao, Ke, Pfister, and Ni]{num16}
Jiancheng Yang, Rui Shi, Donglai Wei, Zequan Liu, Lin Zhao, Bilian Ke, Hanspeter Pfister, and Bingbing Ni.
\newblock Medmnist v2-a large-scale lightweight benchmark for 2d and 3d biomedical image classification.
\newblock \emph{Scientific data}, 10\penalty0 (1):\penalty0 41, 2023.

\bibitem[Zheng et~al.(2025)Zheng, Chen, Gong, Griffin, and Slabaugh]{num6}
Xiaoyu Zheng, Xu~Chen, Shaogang Gong, Xavier Griffin, and Greg Slabaugh.
\newblock Xfmamba: Cross-fusion mamba for multi-view medical image classification.
\newblock In \emph{International Conference on Medical Image Computing and Computer-Assisted Intervention}, pages 672--682. Springer, 2025.

\bibitem[Zhu et~al.(2024{\natexlab{a}})Zhu, Fu, Li, and Larson]{num24}
Qikui Zhu, Chuan Fu, Shuo Li, and K~Larson.
\newblock Class-consistent contrastive learning driven cross-dimensional transformer for 3d medical image classification.
\newblock In \emph{IJCAI}, pages 1807--1815, 2024{\natexlab{a}}.

\bibitem[Zhu et~al.(2024{\natexlab{b}})Zhu, Cai, Wang, Chen, Fu, and Yao]{num20}
Yaoyao Zhu, Xiuding Cai, Xueyao Wang, Xiaoqing Chen, Zhongliang Fu, and Yu~Yao.
\newblock Bsda: Bayesian random semantic data augmentation for medical image classification.
\newblock \emph{Sensors}, 24\penalty0 (23):\penalty0 7511, 2024{\natexlab{b}}.

\end{thebibliography}
\end{document}